\documentclass[a4paper,11pt]{article}
\usepackage{amsmath,amsfonts,graphicx,array,fullpage}
\newcommand{\smf}[2]{\hbox{${#1\over#2}$}}

\newcommand{\pic}[1]{\includegraphics[width=14.5cm]{graphs/#1}}

\newcommand{\M}{\ensuremath\mathcal{M}}
\newcommand{\that}{\hat\theta}
\newcommand{\mhat}{\hat\mu}
\newcommand{\mmax}{{\mu^*}}

\newcommand{\qed}{\hfill\ensuremath{\Box}}
\newcommand{\url}[1]{\texttt{#1}}

\newtheorem{theorem}{Theorem}
\newtheorem{proposition}[theorem]{Proposition}

\newenvironment{proof}[1][Proof]{\begin{trivlist}\item[\hskip \labelsep {\bfseries #1}]}{\qed\end{trivlist}}

\def\KL(#1||#2){D(#1\parallel#2)}

\def\PU_#1(#2){P_{\textnormal{\tiny#1}}\!\left(#2\right)}
\def\LU_#1(#2){L_{\textnormal{\tiny#1}}\!\left(#2\right)}
\def\P(#1|#2){P\!\left(#1\mid#2\right)}
\def\L(#1|#2){L\!\left(#1\mid#2\right)}
\def\PP(#1|#2){P_{\textnormal{\tiny P}}\!\left(#1\mid#2\right)}
\def\PG(#1|#2){P_{\textnormal{\tiny G}}\!\left(#1\mid#2\right)}
\def\LP(#1|#2){L_{\textnormal{\tiny P}}\!\left(#1\mid#2\right)}
\def\LG(#1|#2){L_{\textnormal{\tiny G}}\!\left(#1\mid#2\right)}
\def\IP(#1){I_{\textnormal{\tiny P}}\!\left(#1\right)}
\def\IG(#1){I_{\textnormal{\tiny G}}\!\left(#1\right)}
\def\wp(#1){w_{\textnormal{\tiny P}}\!\left(#1\right)}
\def\wg(#1){w_{\textnormal{\tiny G}}\!\left(#1\right)}

\begin{document}
\title{An Empirical Study of MDL Model Selection with Infinite
  Parametric Complexity}
\author{Steven de Rooij and Peter Gr\"unwald} \maketitle

\begin{abstract}
Parametric complexity is a central concept in MDL model selection. In
practice it often turns out to be infinite, even for quite simple
models such as the Poisson and Geometric families.  In such cases, MDL
model selection as based on NML and Bayesian inference based on
Jeffreys' prior can not be used.  Several ways to resolve this problem
have been proposed. We conduct experiments to compare and evaluate
their behaviour on small sample sizes.

We find interestingly poor behaviour for the plug-in predictive code;
a restricted NML model performs quite well but it is questionable if
the results validate its theoretical motivation. The Bayesian model
with the improper Jeffreys' prior is the most dependable.
\end{abstract}

\section{Introduction}\label{section:introduction}
Model selection is the task of choosing one out of a set of
alternative hypotheses, or models, for some phenomenon, based on the
available data. Let $\M_1$ and $\M_2$ be two parametric models and $D$
be the observed data. Modern MDL model selection proceeds by
associating so-called \emph{universal codes} (formally defined in
Section~\ref{section:universal_codes}) with each model.  We denote the
resulting codelength for the data by $\LU_$\!\M\!$(D)$, which can be
read as `the number of bits needed to encode $D$ with the help of
model $\M$'.  We then pick the model $\M$ that minimises this
expression and thus achieves the best compression of the data.

There exist a variety of universal codes, all of which lead to
different codelengths and thus to different model selection
criteria. While any choice of universal code will lead to an
asymptotically `consistent' model selection method (eventually the
right model will be selected), to get good results for small sample
sizes it is crucial that we select an efficient code. According to the
MDL philosophy, this should be taken to mean a code that compresses
the data as much as possible in a worst-case sense. This is made
precise in Section~\ref{section:universal_codes}. Whenever this code,
known as the Shtarkov or NML code, is well-defined MDL model selection
is straightforward to apply and typically leads to good results 
(see \cite{GrunwaldMP05} and the various experimental papers therein).

However, the worst-case optimal universal code can be ill-defined even
for such simple models as the Poisson and geometric models. In such
cases it is not clear how MDL model selection should be applied. A
variety of remedies to this problem have been proposed, and in this
paper we investigate them empirically, for the simple case where
$\M_1$ is Poisson and $\M_2$ geometric.  We find that some solutions,
most notably the use of a plug-in code instead of the NML code, lead
to relatively poor results.  However, since these codes no longer
minimise the worst-case regret they are harder to justify
theoretically. In fact, as explained in more detail in
Section~\ref{sec:objective}, the only method that may  have an
MDL-type justification closely related to minimax regret is the
Bayesian code with the improper Jeffreys' prior.  Perhaps not
coincidentally, this also seems the most dependable code among all
solutions that we tried.

In Section~\ref{section:universal_codes} we describe the code that
achieves worst-case minimal regret. This code is undefined for the
Poisson and geometric distributions. We analyse these models in more
detail in Section~\ref{section:models}. In Section~\ref{section:ways} we
describe four different approaches to MDL model selection under such
circumstances. We test these using error rate and calibration tests
and evaluate them in Section~\ref{section:results}. Our conclusions
are summarized in Section~\ref{section:conclusion}.

\section{Universal codes}\label{section:universal_codes}

A \emph{universal code} is universal with respect to a set of codes,
in the sense that it codes the data in not many more bits than the
best code in the set, \emph{whichever} data sequence is actually
observed. We thus define universal codes on an individual sequence
basis, as in \cite{BarronRY98}, rather than in an expected
sense. The difference between the codelength of the universal code and
the codelength of the shortest code in the set is called the
\emph{regret}, which is a function of a concrete data sequence, unlike
``redundancy'' which is an expectation value and which we do not use
here.

There exists a one-to-one correspondence between code lengths and
probability distributions: for any probability distribution, a code
can be constructed such that the negative logs of the probabilities
equal the codeword lengths of the outcomes, and vice versa; here we
conveniently ignore rounding issues \cite{Grunwald05}. Therefore we
can phrase our hypothesis testing procedure in terms of
\emph{statistical models}, which are sets of probability
distributions, rather than sets of codes. In this paper, we define
universal codes relative to parametric families of distributions
(`models') $\M$, which we think of as sets of distributions or sets of
code length functions, depending on circumstance. Let $U$ be a code
with length function $L_U$. Relative to a given model $\M$ and sample
$x^n = x_1,\ldots, x_n$, the {\em regret} of $U$ is formally defined
as
\begin{equation}
\label{eq:regretintro}
L_U(x^n) - [ - \ln P(x^n \mid \hat{\theta}(x^n))],
\end{equation}
where $\that(x^n)$ is the maximum likelihood estimator, indexing the
element of $\M$ that assigns maximum probability to the data. It will
sometimes be abbreviated to just $\that$.  Also note that we compute
code lengths in nats rather than bits, this will simplify subsequent
equations somewhat.

The correspondence between codes and distributions referred to above
amounts to the fact we can transform $P_U$ into a corresponding code,
such that the codelengths satisfy, for all $x^n$ in the sample space
${\cal X}^n$,
\begin{displaymath}
L_U(x^n)=-\ln P_U(x^n).
\end{displaymath}
Many different constructions of universal codes have been proposed.
Some are easy to implement, others have nice theoretical properties.
The MDL philosophy \cite{Rissanen96,Grunwald05} has it that the
best universal code minimises the regret (\ref{eq:regretintro}) in the
worst case of all possible data sequences. This ``minimax optimal''
solution is called the ``Normalised Maximum Likelihood'' (NML) code,
which was first described by Shtarkov, who also observed its minimax
optimality properties. The NML-probability of a data sequence for a
parametric model is defined as follows:

\begin{equation}
  \PU_NML(x^n):={\P(x^n|{\that(x^n)})\over\sum_{x^n}\P(x^n|{\that(x^n)})}
\end{equation}
The codelength that corresponds to this probability is called the
\emph{stochastic complexity}. Writing $L(x^n)\equiv-\ln P(x^n)$ we
get:
\begin{equation}
  -\ln\PU_NML(x^n)=\LU_NML(x^n)=\L(x^n|\that)+\ln\sum_{x^n}\P(x^n|{\that(x^n)})\label{eqn:stochastic_complexity}
\end{equation}
The last term in this equation is called the \emph{parametric
complexity}. For this particular universal code, the regret (\ref{eq:regretintro}) is equal
to the parametric complexity for all data sequences, and therefore
constant across sequences of the same length. It is not hard to deduce
that $\PU_NML(\cdot)$ achieves minimax optimal regret with respect to
$\M$. It is usually impossible to compute the parametric complexity
analytically, but it can be shown that, under regularity conditions on
$\M$ and its parametrisation, asymptotically it behaves as follows:
\begin{equation}
  \LU_ANML(x^n) := \L(x^n|\that)+{k\over2}\ln{n\over2\pi}+\ln\int_\Theta\!\!\!\sqrt{\det
  I(\theta)}d\theta\label{eqn:approx}
\end{equation}
Here, $n$, $k$ and $I(\theta)$ denote the number of outcomes, the
number of parameters and the Fisher information matrix
respectively. The difference between $\LU_ANML(x^n)$ and
$\LU_NML(x^n)$ goes to zero as $n\rightarrow\infty$. Since the last
term does not depend on the sample size, it has often been disregarded
and many people came to associate MDL only with the first two
terms. But the third term can be quite large or even infinite, and it
can substantially influence the inference results for small sample
sizes.

Interestingly, Equation~\ref{eqn:approx} also describes the asymptotic
behaviour of the Bayesian universal code where Jeffreys' prior is
used: here MDL and an objective Bayesian method coincide even though
their motivation is quite different.

The parametric complexity can be infinite. Many strategies have been
proposed to deal with this, but most have a somewhat ad-hoc
character. When Rissanen defines stochastic complexity as
(\ref{eqn:stochastic_complexity}) in his 1996 paper \cite{Rissanen96},
he writes that he does so ``thereby concluding a decade long search'',
but as Lanterman observes \cite{Lanterman05}, ``in the light of these
problems we may have to postpone concluding the search just a while
longer.''

\section{The Poisson and geometric models}\label{section:models}
\begin{figure}[t]
\centerline{\includegraphics[width=10cm]{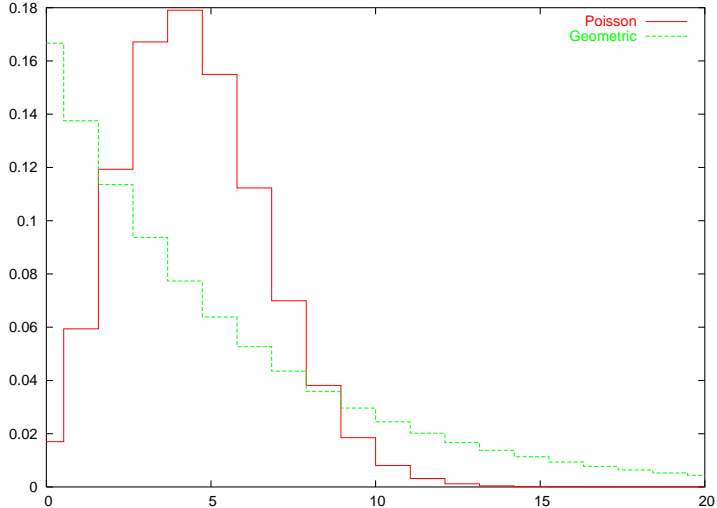}}
\caption{\small\it The Poisson and geometric distributions for
  $\mu=5$.}
\label{fig:distributions}
\end{figure}

We investigate MDL model selection between the Poisson and geometric
models. Figure~\ref{fig:distributions} may help form an intuition
about the probability mass functions of the two distributions. One
reason for our choice of models is that they are both single parameter
models, so that the dominant ${k\over2}\ln{n\over 2\pi}$ term of
Equation~\ref{eqn:approx} cancels. This means that at least for large
sample sizes, picking the model which best fits the data should always
work. We nevertheless observe that for small sample sizes, data which
are generated by the geometric distribution are misclassified as
Poisson much more frequently than the other way around (see
Section~\ref{section:results}). So in an informal sense, even though
the number of parameters is the same, the Poisson distribution is more
prone to `overfitting'. 

To counteract the bias in favour of Poisson that is introduced if we
just select the best fitting model, we would like to compute the third
term of Equation~\ref{eqn:approx}, which now characterises the
parametric complexity. But as it turns out, both models have an
infinite parametric complexity; the integral in the third term of the
approximation also diverges! So in this case it is not immediately
clear how the bias should be removed. This is the second reason why we
chose to study the Poisson and geometric models. In
Section~\ref{section:ways} we describe a number of methods that have
been proposed in the literature as ways to deal with infinite
parametric complexity; in Section~\ref{section:results} they are
evaluated evaluate empirically.

Reassuringly, all methods we investigate tend to `punish' the Poisson
model, and thus compensate for this overfitting phenomenon. However,
the amount by which the Poisson model is punished depends on the
method used, so that different methods give different results.

We parameterise both the Poisson and the Geometric family of
distributions by the mean $\mu\in(0,\infty)$, to allow for easy
comparison. This is possible because for both models, the empirical
mean (average) of the observed data is a sufficient statistic. For
Poisson, parameterisation by the mean is standard. For geometric, the
reparameterisation can be arrived at by noting that in the standard
parameterisation, $P(x \mid \theta) = (1-\theta)^x \theta$, the mean
is given by $\mu = (1-\theta)/\theta$.  As a notational reminder the
parameter is called $\mu$ henceforth. Conveniently, the ML estimator
$\mhat$ for both distributions is the average of the data (the proof
is immediate if we set the derivative to zero).

We will add a subscript {\sc p} or {\sc g} to indicate that
codelengths are computed with respect to the Poisson model or the
geometric model, respectively:

\begin{eqnarray}
  \LP(x^n|\mu)& = &-\ln\prod_{i=1}^n{e^{-\mu}\mu^{x_i}\over
  x_i!}=\sum_{i=1}^n\ln(x_i!)+n\mu-\ln\mu\sum_{i=1}^n x_i\\
  \LG(x^n|\mu)& =
  &-\ln\prod_{i=1}^n{\mu^{x_i}\over(\mu\!+\!1)^{x_i+1}}=n\ln(\mu\!+\!1)-\ln\left({\mu\over\mu\!+\!1}\right)\sum_{i=1}^n x_i
\end{eqnarray}

\section{Four ways to deal with infinite parametric
  complexity}\label{section:ways}
In this section we discuss four general ways to deal with the infinite
parametric complexity of the Poisson and geometric models when the
goal is to do model selection. Each of these four leads to one or
sometimes more concrete selection criteria which we put into practice
and evaluate in Section~\ref{section:results}.

\subsection{BIC/ML}
One way to deal with the diverging term of the approximation is to
just ignore it. The model selection criterion that results corresponds
to only a very rough approximation of any real universal code, but it
has been used and studied extensively. It was proposed both by
Rissanen \cite{Rissanen78} and Schwarz \cite{Schwarz78} in 1978.
Schwarz derived it as an approximation to the Bayesian marginal
likelihood, and for this reason, it is best known as the BIC (Bayesian
Information Criterion). Rissanen already abandoned the idea in the mid
1980's in favour of more sophisticated approximations of the NML
codelength.

\begin{displaymath}
  \LU_BIC(x^n)=\L(x^n|\mhat)+{k\over2}\ln n
\end{displaymath}
Comparing the BIC to the approximated NML codelength we find that in
addition to the diverging term, a ${k\over2}\ln{1\over2\pi}$ term has
also been dropped. This curious difference can be safely ignored in
our setup, where $k$ is equal to one for both models so the whole term
cancels anyway. According to BIC, we
must select the geometric model iff:

\begin{displaymath}
  0 < \LU_P,BIC(x^n)-\LU_G,BIC(x^n)=\LP(x^n|\mhat)-\LG(x^n|\mhat)
\end{displaymath}
We are left with a generalised likelihood ratio test (GLRT). Such a
test has the form ${\PP(x^n|\that)\over\PG(x^n|\that)}{>\atop<}\eta$;
the BIC criterion thus reduces to a GLRT with $\eta=0$, which is also
known as maximum likelihood (ML) testing. It should be expected that
this leads to overfitting and therefore to a bias in favour of the
`more complex' Poisson model.

\subsection{Restricted ANML}\label{section:anml}
One often used method of rescuing the NML approach to MDL model
selection is to restrict the range of values that the parameters can
take to ensure that the third term of Equation~\ref{eqn:approx} stays
finite. 

To compute the approximated parametric complexity of the restricted
models we need to establish the Fisher information first. Using the
formula $I(\theta)=-E_\theta[{d^2\over d\theta^2}\L(x|\theta)]$,
we get:
\begin{eqnarray}
  \IP(\mu)& = &-E_\mu\left[-{x\over\mu^2}\right]={1\over\mu}\\
  \IG(\mu)& = &-E_\mu\left[-{x\over\mu^2}+{x+1\over(\mu+1)^2}\right]={1\over\mu(\mu+1)}
\end{eqnarray}
Now we can compute the last term in the parametric complexity
approximation (\ref{eqn:approx}):

\begin{eqnarray}
  \ln\int_0^\mmax\kern-.4cm\sqrt{\IP(\mu)}d\mu& =
  &\ln\int_0^\mmax\kern-.4cm\mu^{-{1\over2}}d\mu=\ln\left(2\sqrt\mmax\right)\\
  \ln\int_0^\mmax\kern-.4cm\sqrt{\IG(\mu)}d\mu& = &
  \ln\int_0^\mmax\kern-.4cm{1\over\sqrt{\mu(\mu+1)}}d\mu=
  \ln2\ln\left(\sqrt\mmax+\sqrt{\mmax+1}\right)
\end{eqnarray}
The parametric complexities of the restricted models with parameter
ranges $(0,\mmax)$ are both monotonically increasing functions of
$\mmax$. However, the parametric complexity of the restricted Poisson
model grows \emph{faster} with $\mmax$ than the parametric complexity
of the geometric model, indicating that the Poisson model has more
descriptive power, even though the models have the same number of
parameters. Let the function $\delta(\mmax)$ measure the difference
between the parametric complexities. Interestingly, this function is
still monotonically increasing in $\mmax$ (this is proven in
Appendix~\ref{appendix:proofs}). It is also easy to prove that it grows
unboundedly in $\mmax$.

\subsubsection{Basic restricted ANML}
We have experimented with restricted models where the parameter range
was restricted to $(0,\mmax)$ for $\mmax\in\{10,100,1000\}$.

This means that we obtain a model selection criterion that selects the
geometric model iff
$0<\LU_{P,ANML$(\mmax)$}(x^n)-\LU_{G,ANML$(\mmax)$}(x^n)=\LP(x^n|\mhat)-\LG(x^n|\mhat)+\delta(\mmax)$.
This is equivalent to a GLRT with threshold $\delta(\mmax)$. For
$\lim\mmax\downarrow0$ we obtain the BIC/ML selection criterion;
higher values of $\mmax$ translate to a selection threshold more in
favour of the geometric model.

An obvious conceptual problem with the resulting code is that the
imposed restriction is quite arbitrary and requires a priori knowledge
about the generating process. But the parameter range can be
interpreted as a hyper-parameter, which can be incorporated into the
code using several techniques, some of which we will discuss.

\subsubsection{Two-part ANML}
The most obvious way to generalise the restricted ANML codelength is
to separately encode a parameter range using some suitable
discretisation. For a sequence with empirical mean $\mhat$, we encode
the integer $b=\left\lceil\log_2\mhat\right\rceil$.  After outputting
$b$ the code length for the rest of the data is computed using
restricted ANML with range $(2^{b-1}, 2^b]$. By taking the logarithm
we ensure that the number of bits used in coding the parameter range
grows at a negligible rate compared to the code length of the data
itself, but we admit that the code for the parameter range admits of
much greater sophistication. We do not really have reason to assume
that the optimal discretisation should be the same for the Poisson and
geometric models for example.

The two-part code is slightly redundant, since code words are assigned
to data sequences of which the ML estimator lies outside the range
that was encoded in the first part -- such data sequences cannot
occur, since for such a sequence we would have encoded a different
range. Furthermore, the two-part code is no longer minimax optimal, so
it is no longer clear why it should be better than other universal
codes which are not minimax optimal. However, as argued in
\cite{Grunwald05}, whenever the minimax optimal code is not defined,
we should aim for a code $L$ which is `close' to minimax optimal in
the sense that, for any compact subset ${\cal M}'$ of the parameter
space, the additional regret of $L$ on top of the NML code for ${\cal M}'$ 
should be
small, e.g. $O(\log \log n)$. The two-part ANML code is one of many
universal codes satisfying this `almost minimax optimality'. While it
may not be better than another almost minimax optimal universal code,
it certainly is better than universal codes which do not have the
almost minimax optimality property.

\subsubsection{Renormalised Maximum Likelihood}
Related to the two-part restricted ANML, but more elegant, is
Rissanen's {\em renormalised maximum likelihood\/} (RNML) code,
\cite{Rissanen00,Grunwald05}. This is perhaps the most widely known
approach to deal with infinite parametric complexity. The idea here
is, rather than explicitly encoding a parameter range in which we know
the ML estimator to lie, to treat the range $(0,\mmax)$ as a
hyper-parameter. The probability of a sequence is computed using the
maximum likelihood value for the hyper-parameter, but just as ordinary
NML has to be normalised, the probability distribution thus obtained
must be \emph{renormalised} to ensure that it sums to $1$.

If this still leads to infinite parametric complexity, we define a
hyper-hyper-parameter. We repeat the procedure until the resulting
complexity is finite. Unfortunately, in our case, after the first
renormalisation, both parametric complexities are still infinite; we
did not manage to perform a second renormalisation. Therefore, we have
not experimented with the RNML code.

\subsection{Plug-in predictive code}
The \emph{plug-in predictive code}, or prequential ML code, is an
attractive universal code because it is usually a lot easier to
implement than either NML or a Bayesian code. Moreover, its
implementation hardly requires any arbitrary decisions. Here the
outcomes are coded sequentially using the probability distribution
indexed by the ML estimator for the previous outcomes
\cite{Dawid84,Rissanen84}; for a general introduction see
\cite{Grunwald05}.
\begin{displaymath}
  \LU_PIPC(x^n)=\sum_{i=1}^n \L(x_i|{\mhat(x^{i-1})})
\end{displaymath}
where $\L(x_i |{\mhat(x^{i-1})}) = - \ln\P(x_i | {\mhat(x^{i-1})})$ is
the number of nats needed to encode outcome $x_i$ using the code based
on the ML estimator on $x^{i-1}$. We further discuss the motivation
for this code in Section~\ref{sec:poor}.

For both the Poisson model and the geometric model, the maximum
likelihood estimator is not well-defined until after a nonzero outcome
has been observed (since $0$ is not inside the allowed parameter
range). This means that we need to use another code for the first few
outcomes. It is not really clear how we can use the model assumption
(Poisson or geometric) here, so we pick a simple code for the
nonnegative integers that does not depend on the model. This will
result in the same codelength for both models; therefore it does not
influence which model is selected. Since there are usually only very
few initial zero outcomes, we may reasonably hope that the results
are not too distorted by our way of handling this startup problem.
We note that this startup problem is an inherent feature of the
predictive plug-in approach \cite{Dawid84,Rissanen89}, and our way of
handling it is in line with the suggestions in \cite{Dawid84}.

\subsection{Objective Bayesian approach}
\label{sec:objective}
In the Bayesian framework we select a prior $w(\theta)$ on the unknown
parameter and compute the marginal likelihood
$\PU_BAYES(x^n)=\int_\Theta \P(x^n|\theta)w(\theta)d\theta$, which
corresponds to a universal code $\LU_BAYES(x^n) = - \ln \PU_BAYES(x^n)$.
Like the NML, this can be
approximated with an asymptotic formula. For exponential families such
as the models under consideration, we have \cite{Balasubramanian97}:
\begin{equation}\label{eqn:bayesapprox}
  \LU_ABAYES(x^n):=\L(x^n|\that)+{k\over2}\ln{n\over2\pi}+\ln{\sqrt{\det
  I(\theta)}\over w(\theta)},
\end{equation}
where
$\LU_ABAYES(x^n) - \LU_BAYES(x^n) \rightarrow 0$ as $n \rightarrow \infty$.
Objective Bayesian reasoning suggests we use Jeffreys' prior
\cite{BernardoSmith1994} for several reasons; one reason is that it is
uniform over all `distinguishable' elements of the model, which
implies that the obtained results are independent of the
parametrisation of the model \cite{Balasubramanian97}. It is defined
as follows:
\begin{equation}
  w(\theta)={\sqrt{\det I(\theta)}\over\int_\Theta\sqrt{\det I(\theta)}d\theta}
\end{equation}
Unfortunately, the normalisation factor in Jeffreys' prior diverges
for both the Poisson model and the geometric model. But if one is
willing to accept a so-called \emph{improper} prior, which is not
normalised, then it is possible to compute a perfectly proper Bayesian
posterior, after observing the first outcome, and use that as a prior
to compute the marginal likelihood of the rest of the data.  The
resulting universal codes with lengths $\LU_BAYES(x_2, \ldots, x_n
\mid x_1)$ are, in fact, {\em conditional\/} on the first
outcome. Recent work by Li and Barron \cite{LiangB05} suggests that,
at least asymptotically and for one-parameter models, the universal
code achieving the minimal {\em expected redundancy conditioned on the
first outcome\/} is given by the Bayesian universal code with the
improper Jeffreys' prior. Li and Barron only prove this for scale and
location models, but their result at least suggests that the same
would still hold for general exponential families such as Poisson and
geometric. It is possible to define MDL inference in terms of either
the expected redundancy or of the worst-case regret. In fact, the
resulting procedures are very similar, see \cite{BarronRY98}. Thus,
we have a tentative justification for using Jeffreys' prior also from
an MDL point of view, on top of its justification in terms of
objective Bayes.

To make this idea more concrete, we compute Jeffreys' posterior after
observing one outcome, and use it to find the Bayesian marginal
likelihoods. We write $x_i^j$ to denote $x_i,\ldots,x_j$ and
$\mhat(x_i^j)$ to indicate which outcomes determine the ML estimator,
finally we abbreviate $s_n=x_1+\ldots+x_n$. The goal is to compute
$\PU_BAYES(x_2^n\mid x_1)$ for the Poisson and geometric models. As
before, the difference between the corresponding codelengths defines a
model selection criterion. We also compute $\PU_ABAYES(x_2^n\mid x_1)$
for both models, the approximated version of the same quantity, based
on approximation formula~(\ref{eqn:bayesapprox}). Equations for the
Poisson and geometric models are presented below.

\paragraph{Bayesian code for the Poisson model}
We compute Jeffreys' improper prior and the posterior after observing
one outcome:
\begin{eqnarray}
  \wp(\mu)& \propto &\sqrt{\IP(\mu)}=\mu^{-{1\over2}}\\
  \wp(\mu\mid x_1)& =
  &{\PP(x_1|\mu)\wp(\mu)\over\int_0^\infty\PP(x_1|\theta)\wp(\theta)d\theta}
  ={e^{-\mu}\mu^{x_1-{1\over2}}\over\Gamma(x_1+{1\over2})}
\end{eqnarray}
From this we can derive the marginal likelihood of the rest of the
data. The details of the computation are omitted for brevity.
\begin{eqnarray}
  \PU_P,BAYES(x_2^n\mid x_1)& =
  &\int_0^\infty\kern-.3cm\PP(x_2^n|\mu)\wp(\mu\mid
  x_1)d\mu\nonumber\\
  & = &{\Gamma(s_n+{1\over2})\over\Gamma(x_1+{1\over2})}/\left(n^{s+{1\over2}}\prod_{i=2}^n x_i!\right)
\end{eqnarray}
We also compute the approximation for the Poisson model using
(\ref{eqn:bayesapprox}):
\begin{multline}
  \LU_P,ABAYES(x_2^n\mid x_1)=\LP(x_2^n|\mhat(x_2^n))+\smf{1}{2}\ln{n\over2\pi}+\ln{\sqrt{\IP(\mhat(x_2^n))}\over\wp(\mhat(x_2^n)\mid x_1)}\\
  =\LP(x_2^n|\mhat(x_2^n))+\smf{1}{2}\ln{n\over2\pi}+\mhat(x_2^n)-x_1\ln\mhat(x_2^n)+\ln\Gamma(x_1+\smf{1}{2})
\end{multline}

\paragraph{Bayesian code for the geometric model}
We perform the same computations for the geometric model. Jeffreys'
improper prior and its posterior after one outcome are:
\begin{eqnarray}
  \wg(\mu)& \propto &\mu^{-{1\over2}}(\mu+1)^{-{1\over2}}\\
  \wg(\mu\mid x_1)& = &(x_1+\smf{1}{2})\mu^{x_1-{1\over2}}(\mu+1)^{-x_1-{3\over2}}\\
  \PU_G,BAYES(x^n)& = &(x_1+\smf{1}{2}){\Gamma(s+{1\over2})\Gamma(n)\over\Gamma(n+s+{1\over2})}
\end{eqnarray}
For the approximation we obtain:
\begin{multline}
  \LU_G,ABAYES(x_2^n\mid
  x_1)=\LG(x_2^n|\mhat(x_2^n))+{1\over2}\ln{n\over2\pi}+\\
  x_1\ln\left(1+{1\over\mhat(x_2^n)}\right)+\smf{1}{2}\ln(\mhat(x_2^n))-\ln(x_1+\smf{1}{2})
\end{multline}

\section{Results}\label{section:results}
We have now described how to compute or approximate the length of a
number of different universal codes, which can be used in an MDL model
selection framework. The MDL principle tells us to select the model
using which we can achieve the shortest codelength for the data. This
coincides with the Bayesian maximum a-posteriori (MAP) model with a
uniform prior on the models. In this way each method for computing or
approximating universal codelengths defines a model selection
criterion, which we want to compare empirically.

In addition to the criteria that are based on universal codes, as
developed in Section~\ref{section:ways}, we define one additional,
`ideal' criterion to serve as a reference by which the others can be
evaluated. The \emph{known $\mu$} criterion cheats a little bit: it
computes the code length for the data with knowledge of the mean of
the generating distribution. If the mean is $\mu$, then the known
$\mu$ criterion selects the Poisson model if
$\LP(x^n|\mu)<\LG(x^n|\mu)$. Since this criterion uses extra knowledge
about the data, it should be expected to perform better than the other
criteria.

We perform two types of test on the selection criteria:

\begin{itemize}
  \item
    Error probability measurements. Here we select a mean and a
    generating distribution (Poisson or geometric), then artificially
    generate samples with that mean of varying size. We offset the
    sample size to the observed probabilities of error of the
    selection criteria. This test can be generalised by putting a prior
    on the generating distribution, such that a fixed percentage of
    samples is generated using a geometric distribution, and the
    remainder using Poisson.

  \item
    Calibration testing. In the Bayesian framework the criteria do not
    only output which model is most likely (the MAP model), but each
    model is also assigned a probability. This can also be understood
    from an MDL perspective via the correspondence between code
    lengths and probabilities we mentioned earlier. If a well
    calibrated selection criterion assigns a probability $p$ to the
    data being Poisson distributed, and the criterion is used
    repeatedly on independently generated data sets, then the
    frequency with which it actually is Poisson should be close to
    $p$. To test this we discretise the probability that each
    criterion assigns to the data being Poisson into a number of
    `bins'. Then we plot for each bin the fraction of samples that
    actually were Poisson.
\end{itemize}
Roughly, the results of both tests can be summarized as follows:
\begin{itemize}

\item
  As was to be be expected, the known $\mu$ criterion performs
  excellently on both tests.

\item
  The criteria based on the plug-in predictive code and BIC/ML
  exhibit the worst performance.

\item
  The basic restricted ANML criterion yields results that range from
  good to very bad, depending on the chosen parameter range. Since the
  range must be chosen without any additional knowledge of the
  properties of the data, this criterion is a bit arbitrary.

\item
  The results for the two-part restricted ANML and Objective Bayesian
  criteria are reasonable in all tests we performed; these criteria
  thus display robustness.
\end{itemize}
In subsequent sections we will treat the results in more detail. In
Section~\ref{section:perr} we discuss the results of the error
probability tests, the calibration test results are presented in
Section~\ref{section:cali}.

\begin{figure}[t]
\begin{center}
  \pic{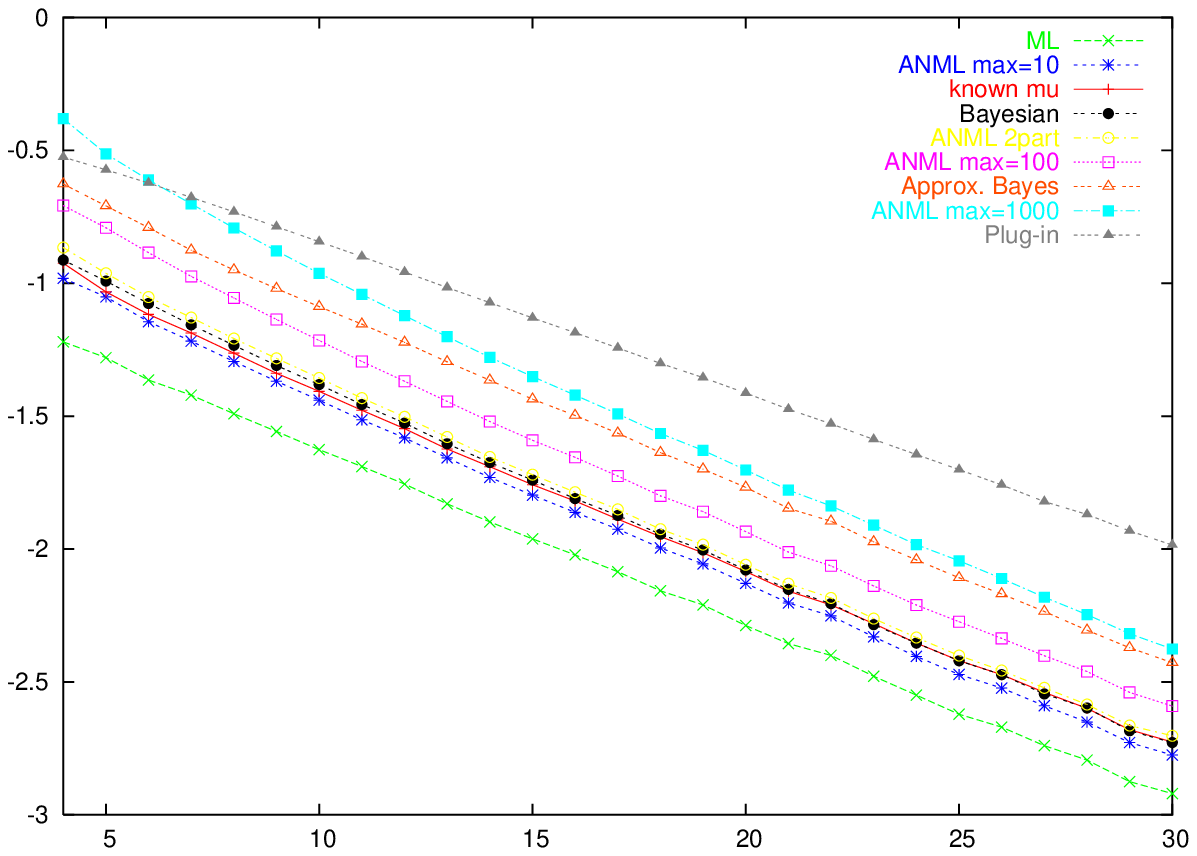}
  \pic{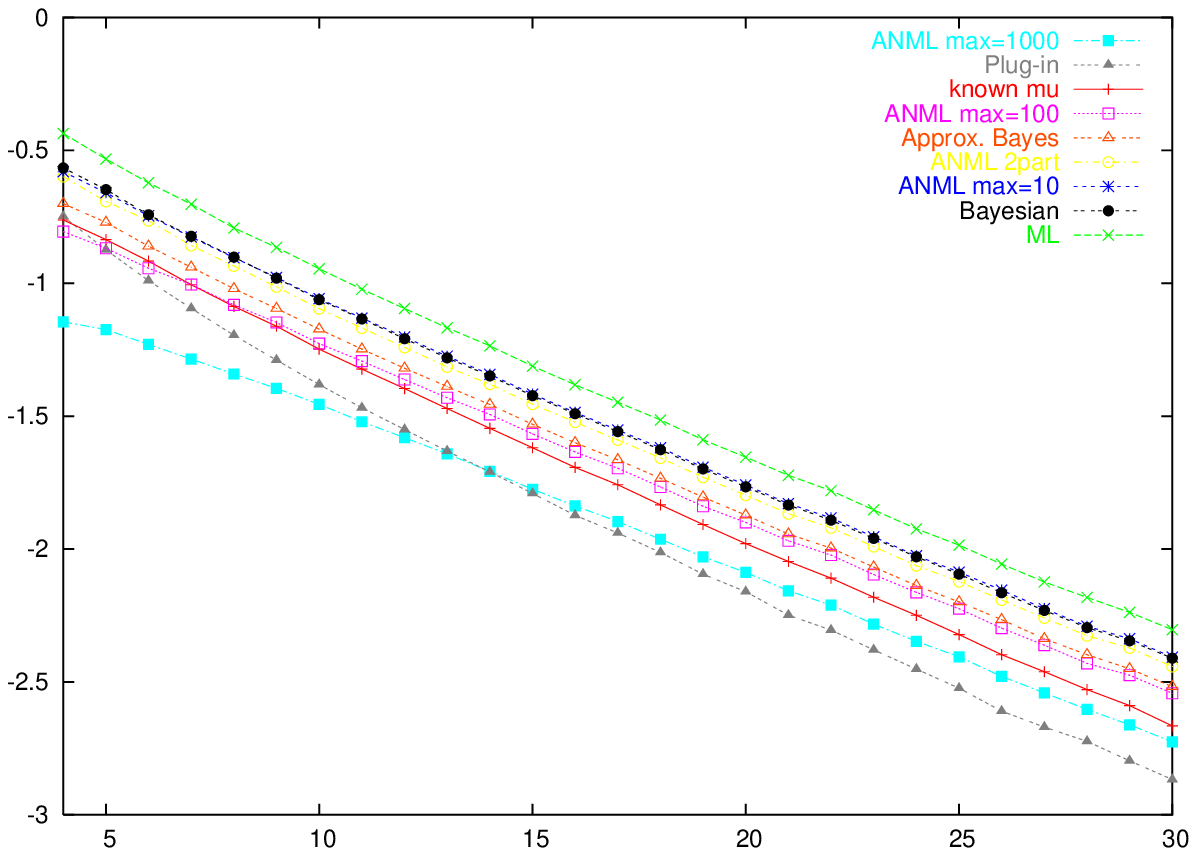}
\end{center}
\caption{\small The $\log_{10}$ of the frequency of wrong
  classification when the data were generated with mean $4$ by the
  Poisson (top) or geometric (bottom) distributions.}
\label{fig:mean4-1}
\end{figure}

\subsection{Error probability}\label{section:perr}
\begin{figure}[t]
\begin{center}
  \pic{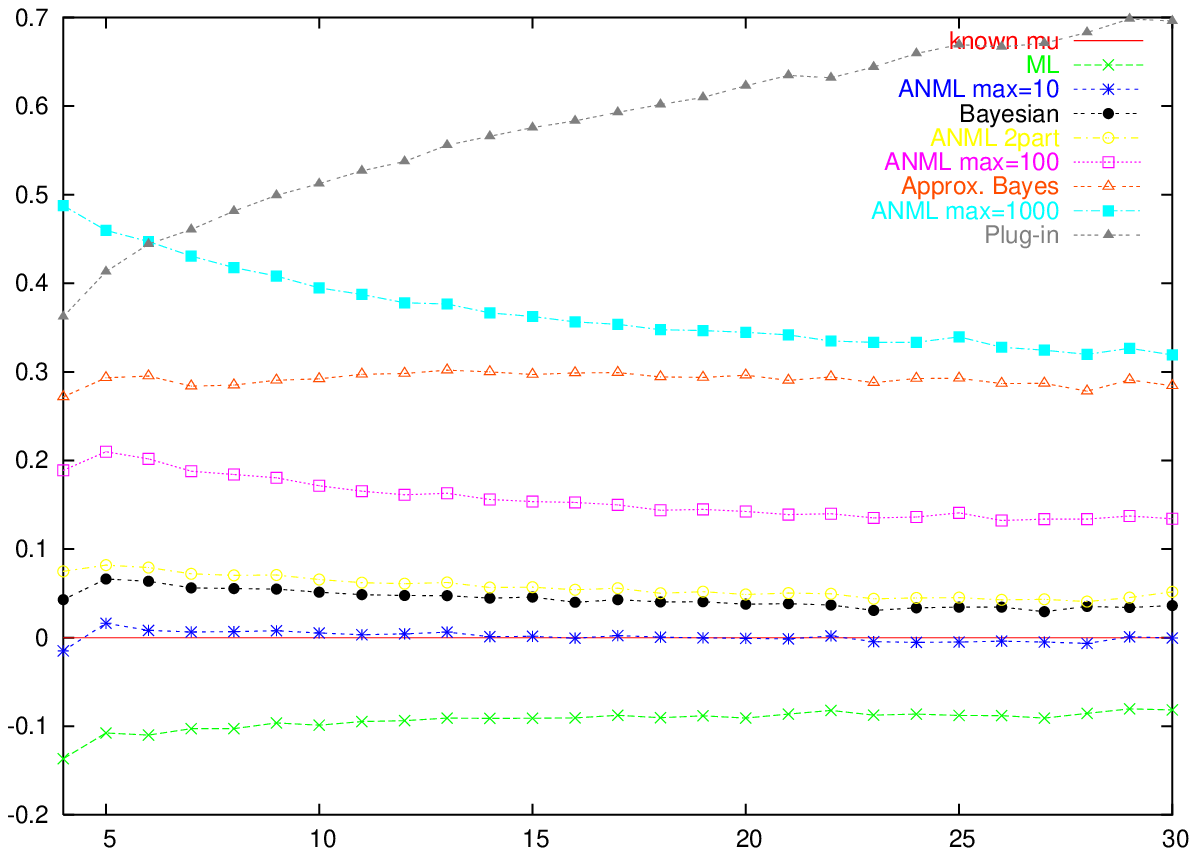}
  \pic{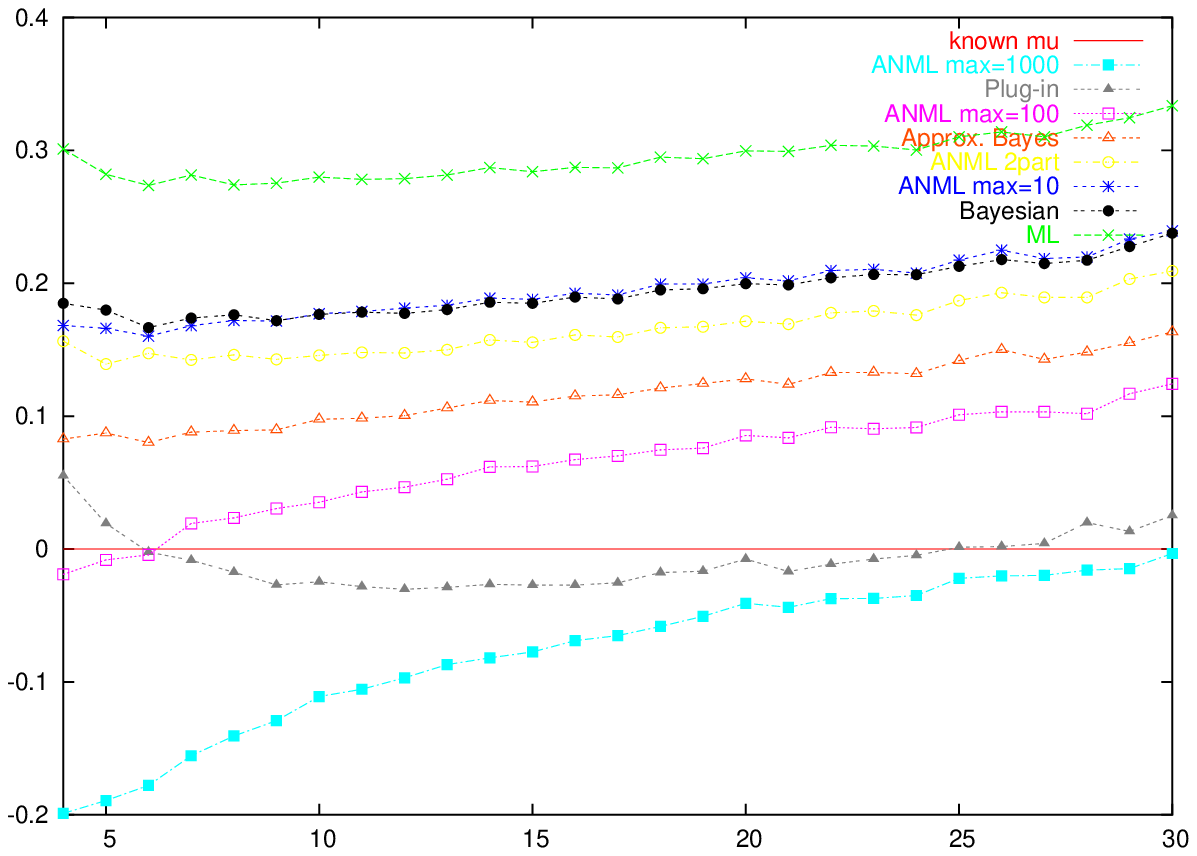}
\end{center}
\caption{\small The difference in the $\log_{10}$ of the frequency of
  wrong classification between each criterion and the known $\mu$
  criterion. In the top graph $90\%$ of the data were generated by
  Poisson, in the bottom graph $10\%$. The mean is $4$.}
\label{fig:mean4-2}
\end{figure}

\begin{figure}[t]
\begin{center}
  \pic{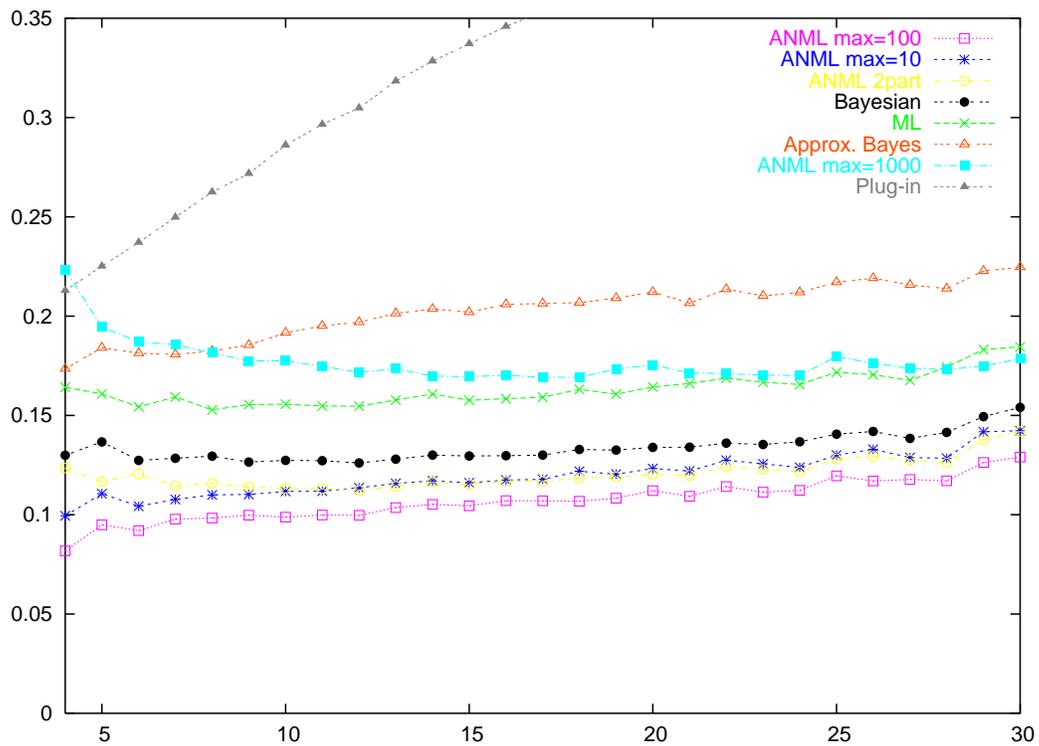}
\end{center}
\caption{\small The difference in the $\log_{10}$ of the frequency of
  wrong classification between each criterion and the known $\mu$
  criterion. Half the data were generated by the Poisson distribution,
  and half by the geometric distribution. The mean is $4$.}
\label{fig:mean4-3}
\end{figure}

In this section we investigate the probability of error of the model
selection criteria through various experiments. We investigate the
error probability in more than one way. First we just look at its
magnitude -- the lower the error probability of a criterion, the
better, obviously. However the error probability in itself is only
defined given a prior on the generating distribution, because the
error probability may differ, depending on whether the data are
Poisson or geometrically distributed. Figures~\ref{fig:mean4-1},
\ref{fig:mean4-2}, \ref{fig:mean4-3} and \ref{fig:mean8,16} show
results of this kind, using different means and different priors on
the generating distribution.

\begin{figure}[t]
\begin{center}
  \pic{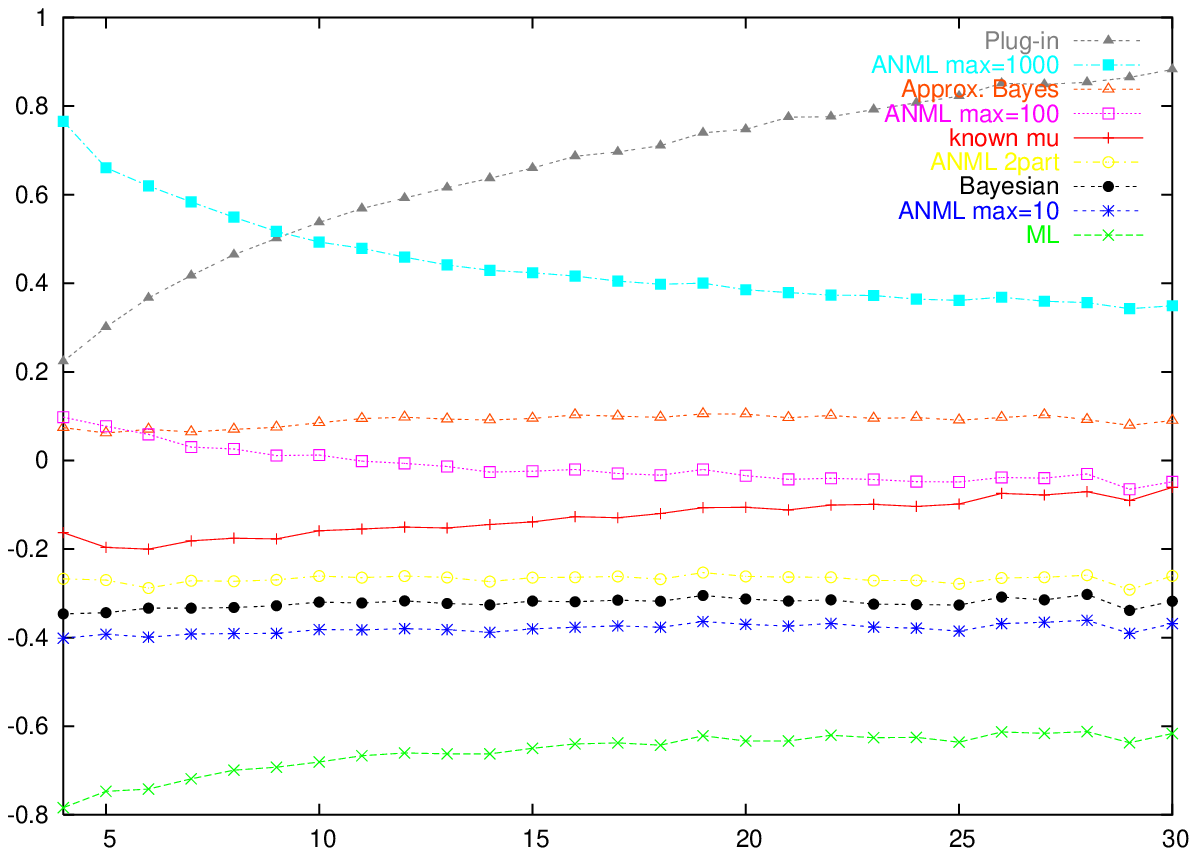}
  \pic{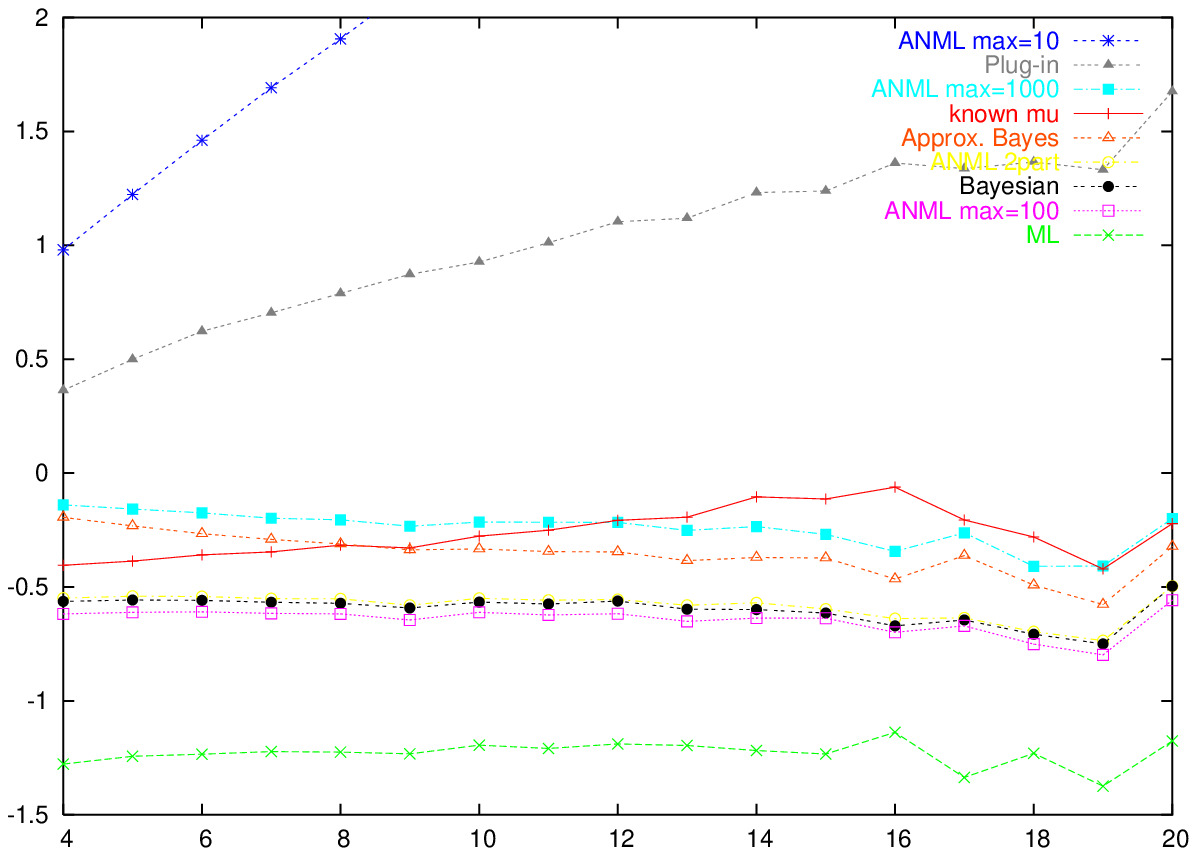}
\end{center}
\caption{\small The classification bias (using log base $10$) when the
  data were generated with mean $4$ (top graph) and mean $16$ (bottom
  graph).}
\label{fig:bias}
\end{figure}

The other interesting aspect of error probability is the notion of
bias: some criteria are much more likely to misclassify Poisson data
than geometric data, or vice versa. By \emph{bias} we mean the
difference between the logs of the error probability for the Poisson
and geometric data. This is not the standard statistical definition
(which is defined with respect to an estimator), but it is a
convenient quantification of our intuitive notion of bias in this
setting. A small bias is an important feature of a good model
selection criterion. Figure~\ref{fig:bias} shows the bias of the
selection criteria; the top graph simply measures the distance between
corresponding lines in the top and bottom graphs of
Figure~\ref{fig:mean4-1}. It is important to realize that a criterion
may be terribly biased and yet have a very low probability of either
type of error -- this is why it is necessary to study both aspects. We
will say more about bias in our discussion of the individual selection
criteria.

\begin{figure}[t]
\begin{center}
  \pic{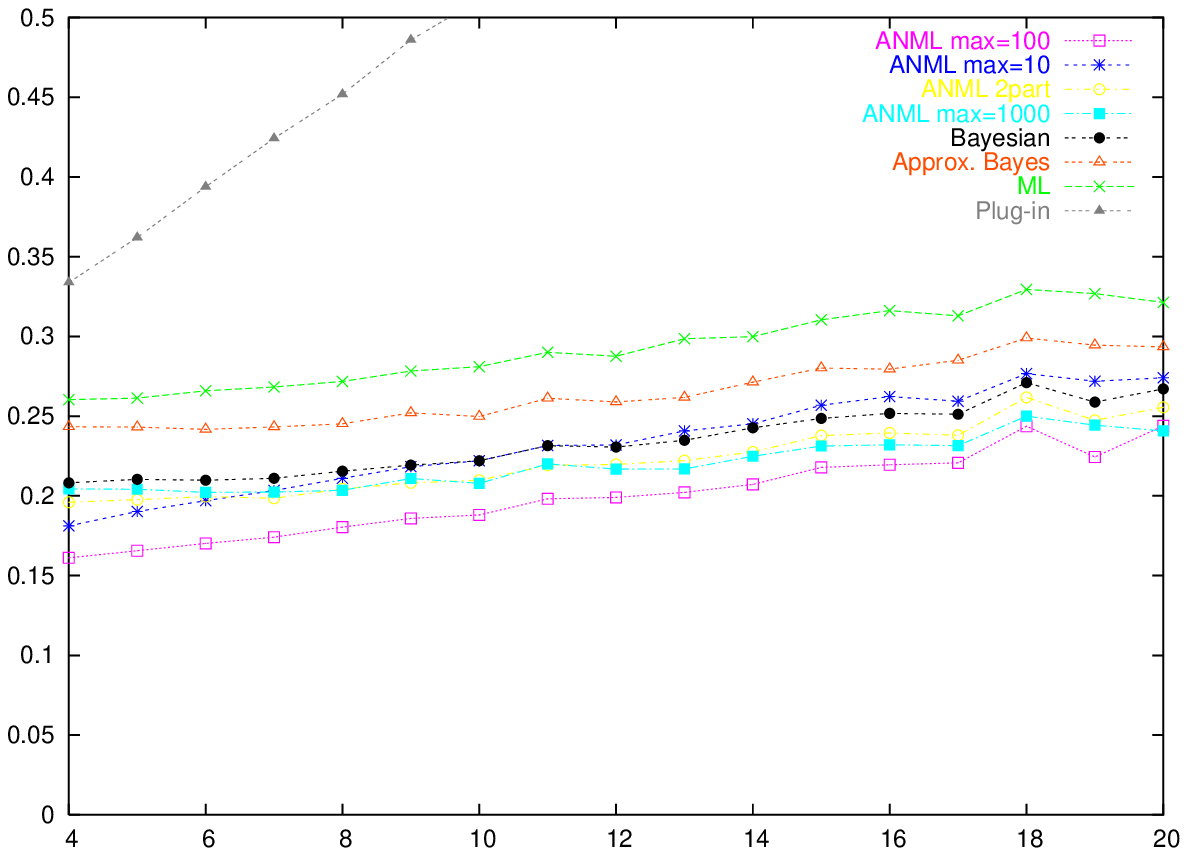}
  \pic{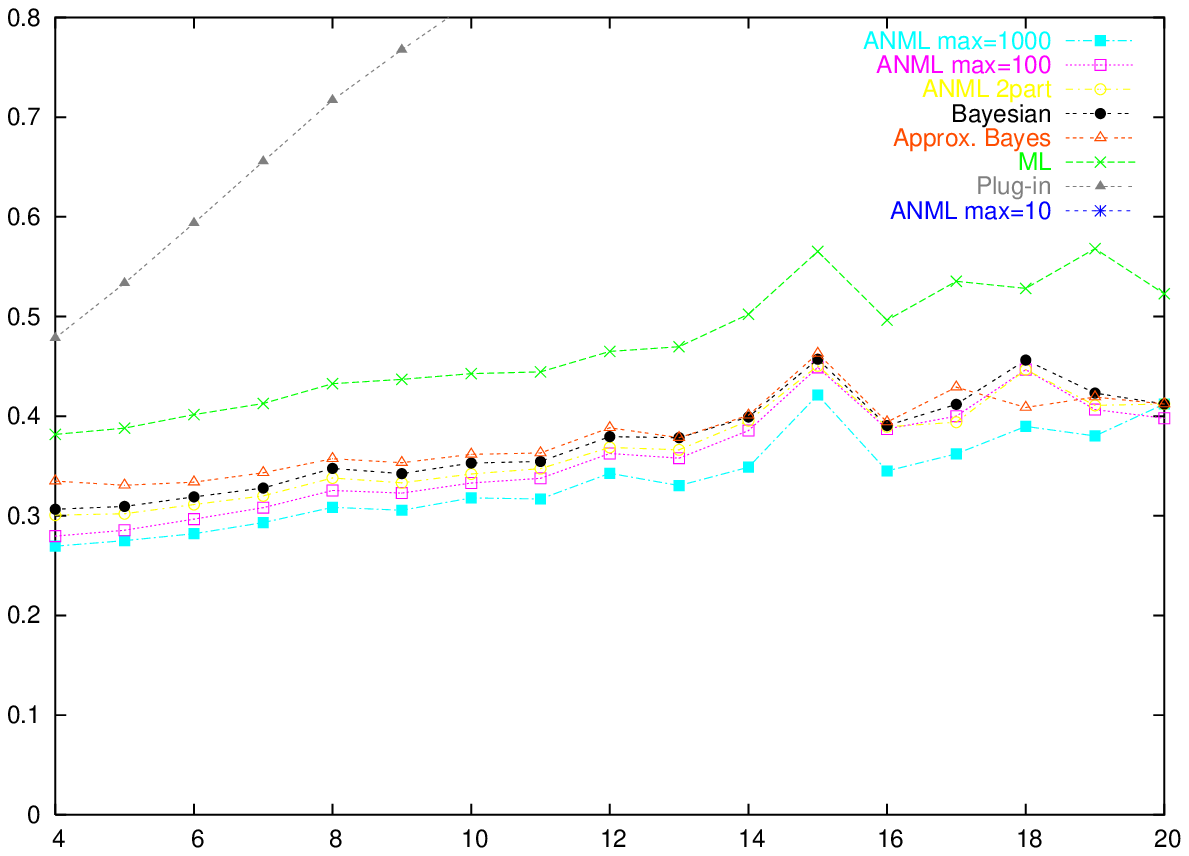}
\end{center}
\caption{\small The difference in the $\log_{10}$ of the frequency of
  wrong classification between each criterion and the known $\mu$
  criterion. Half the data were generated by the Poisson distribution,
  and half by the geometric distribution. The mean is $8$ in the top
  graph and $16$ in the bottom graph.}
\label{fig:mean8,16}
\end{figure}

\subsubsection{Known $\mu$}\label{section:known_mu}
The theoretical analysis of the known $\mu$ criterion is helped by the
circumstance that (1) one of the two hypotheses equals the generating
distribution and (2) the sample consists of outcomes which are
i.i.d. according to this distribution. Cover and Thomas
\cite{CoverT91} use Sanov's~Theorem to show that in such a situation,
the probability of error decreases exponentially in the sample
size. If the Bayesian MAP model selection criterion is used then the
following happens: if the data are generated using
$\textnormal{Poisson}[\mu]$ then the probability of error decreases
exponentially in the sample size, with some error exponent; if the
data are generated with $\textnormal{Geom}[\mu]$ then the overall
probability is exponentially decreasing with the same exponent
\cite[Theorem 12.9.1 on page 312 and text thereafter]{CoverT91}.
Thus, we expect that the line for the ``known $\mu$'' criterion in
Figure~\ref{fig:mean4-1} is straight on a logarithmic scale, with a
slope that is equal on the top and bottom graphs. This proves to be
the case.

\medskip\noindent In Figures~\ref{fig:mean4-2}, \ref{fig:mean4-3} and
\ref{fig:mean8,16} the log of the error frequency of the known $\mu$
criterion is subtracted from the logs of the error frequencies of the
other criteria. This brings out the differences in performance in more
detail. The known $\mu$ criterion, which is perfectly calibrated (as
we will observe later) and which also has a low probability of error
under all circumstances (although biased criteria can sometimes do
better if the bias is in the right direction), is thus treated as a
baseline of sorts.

\subsubsection{Poor performance of the plug-in criterion}
\label{sec:poor}
One feature of Figure~\ref{fig:mean4-1} that immediately attracts
attention is the unusual slope of the error rate line of the plug-in
criterion, which clearly favours the geometric distribution.
We did not obtain completely unbiased results in any of our
experiments, but only the bias of the plug-in criterion grows
substantially as the sample size increases. Figure~\ref{fig:bias}
shows how at $n=30$ its bias in favour of the geometric model reaches
a level where a sample is about $10^{0.88}\approx7$ times more likely
to be misclassified as geometric than as Poisson.

While this behaviour may seem appropriate if the data are judged more
likely to come from a geometric distribution, there is actually a
strong argument that \emph{even under those circumstances it is
not}. As pointed out in the beginning of this section, beside the bias
also the magnitude of the error probability should be taken into
account. Suppose that we put a fixed prior on the generating
distribution, with nonzero probability for both distributions. The
marginal probability of error is a linear combination of the
probabilities of error for the two generating distributions; as such
it is dominated by the probability of error with the \emph{worst}
exponent. So if minimising the probability of error is our goal, then
we must conclude that the behaviour of the plug-in criterion is
suboptimal. (On a side note, minimising the probability of error with
respect to a fixed prior is \emph{not} the goal of classical
hypothesis testing, since in that setting the two hypotheses do not
play a symmetrical role.) To illustrate, the bottom graph in
Figure~\ref{fig:mean4-2} shows that, even if there is only a $10\%$
chance that the data are Poisson, then the plug-in criterion still has
a worse (marginal) probability of error than ``known $\mu$'' as soon
as the sample size reaches 25. Figure~\ref{fig:mean4-3} shows what
happens if the prior on the generating distribution is uniform --
using the plug-in criterion immediately yields the largest probability
of error of all the criteria under consideration. This effect only
becomes stronger if the mean is higher.

This strangely poor behaviour of the plug-in criterion initially came
as a complete surprise to us. Theoretical literature certainly had not
suggested it. Rather the contrary: in 1989 Rissanen writes that ``it
is only because of a certain inherent singularity in the process [of
plug-in coding], as well as the somewhat restrictive requirement that
the data must be ordered, that we do not consider the resulting
predictive code length to provide another competing definition for the
stochastic complexity, but rather regard it as an
approximation.'' \cite{Rissanen89} There are also many results
showing that the regret for the plug-in code grows as ${k\over2}\ln
n$, the same as the regret for the NML code, for a variety of
models. Examples are \cite{Rissanen1986c,Gerencser1987,Wei1990}. So we
were extremely puzzled by these results at first.

To gain intuition as to why the plug-in code should behave so
strangely, note that the variance of a geometric distribution is much
larger than the variance of the Poisson distribution with the same
mean. This suggests that the penalty for using $\hat\mu$ rather than
$\mu$ to code each consecutive outcome is higher for the Poisson
model. The accumulated difference accounts for the difference in
regret.

We have made this intuition precise in a separate publication. In
\cite{GrunwaldR05} we prove that for single parameter exponential
families, the regret for the plug-in code grows with
${1\over2}\ln(n)\textnormal{Var}_P(X)/\textnormal{Var}_M(X)$, where $n$ is
the sample size, $P$ is the generating distribution and $M$ is the
best element of the model (the element of $\M$ for which $\KL(P||M)$
is minimised). The plug-in model has the same regret (to $O(1)$) as
the NML model if and only if the variance of the generating
distribution is the same as the variance of the best element of the
model. The existing literature studies the case where $M=P$, so
automatically $\textnormal{Var}_M(X)=\textnormal{Var}_P(X)$.

\subsubsection{ML/BIC}
Beside known $\mu$ and plug-in, all criteria seem to share more or
less the same error exponent. Nevertheless, they still show
differences in bias. While we have to be careful to avoid
over-interpreting our results, we find that the ML criterion
consistently displays the largest bias in favour of the Poisson
model. Figure~\ref{fig:bias} shows how the ML criterion misclassifies
a sequence as Poisson about four times more often than the other way
around when the mean is $4$; when the mean is raised to $16$ bias has
increased even further, to a ratio of twenty to one.

This illustrates how the Poisson model appears to have a greater
descriptive power, even though the two models have the same number of
parameters, an observation which we hinted at in
Section~\ref{section:models}. Intuitively, the Poisson model allows
more information about the data to be stored in the parameter
estimate. All the other selection criteria compensate for this effect,
by giving a higher probability to the geometric model. (In terms of
coding, the Poisson codelength is increased by more than the geometric
codelength.) Comparing the two graphs in Figure~\ref{fig:bias}, we
find that as the mean of the generating distribution is increased, the
prediction errors for all criteria except ML move closer together,
showing that for higher means it becomes even more worthwhile to try
and compensate for the favouritism of ML/BIC.

\subsubsection{Basic restricted ANML}
We have seen that the ML/BIC criterion shows the largest bias for the
Poisson model. The top graph of Figure~\ref{fig:bias} shows that the
largest bias in the other direction is achieved by ANML $\mmax=1000$
(until the sample gets so large that, inevitably, it is overtaken by
the plug-in criterion).  Apparently the ANML criterion
overcompensates. In a way it is obvious that we could obtain such a
result since we observed in Section~\ref{section:anml} that ANML leads
to a selection criterion that is equivalent to a GLRT with a selection
threshold that is an unbounded, monotonically increasing function of
$\mmax$. Essentially, by choosing an appropriate $\mmax$ we can get
\emph{any} bias in favour of the geometric model. We conclude that it
does not really make sense to use an arbitrarily chosen restricted
parameter domain to repair the NML model when it is undefined.

\subsubsection{Objective Bayes and two-part restricted ANML}
We will not try to interpret the differences in error probability for
the (approximated) Bayesian and ANML 2-part criteria. Since we are
using different selection criteria we should expect at least some
differences in the results. These differences are exaggerated by our
setup with its low mean and small sample size.

Figures~\ref{fig:mean4-2}--\ref{fig:mean8,16} show that the
probability of error for these criteria tends to decrease at a
slightly lower rate than for known $\mu$ (except when the prior on the
generating distribution is heavily in favour of Poisson). While we do
not understand this phenomenon well enough so as to prove it
mathematically, it is of course consistent with the general rule that
with more prior uncertainty, more data are needed to make the right
decision.  It may be that all the information contained within a
sample can be used to improve the resolution of the known $\mu$
criterion, while for the other criteria some of that information has
to be sacrificed in order to estimate the parameter value.

\begin{figure}[!t]
\begin{center}
  \pic{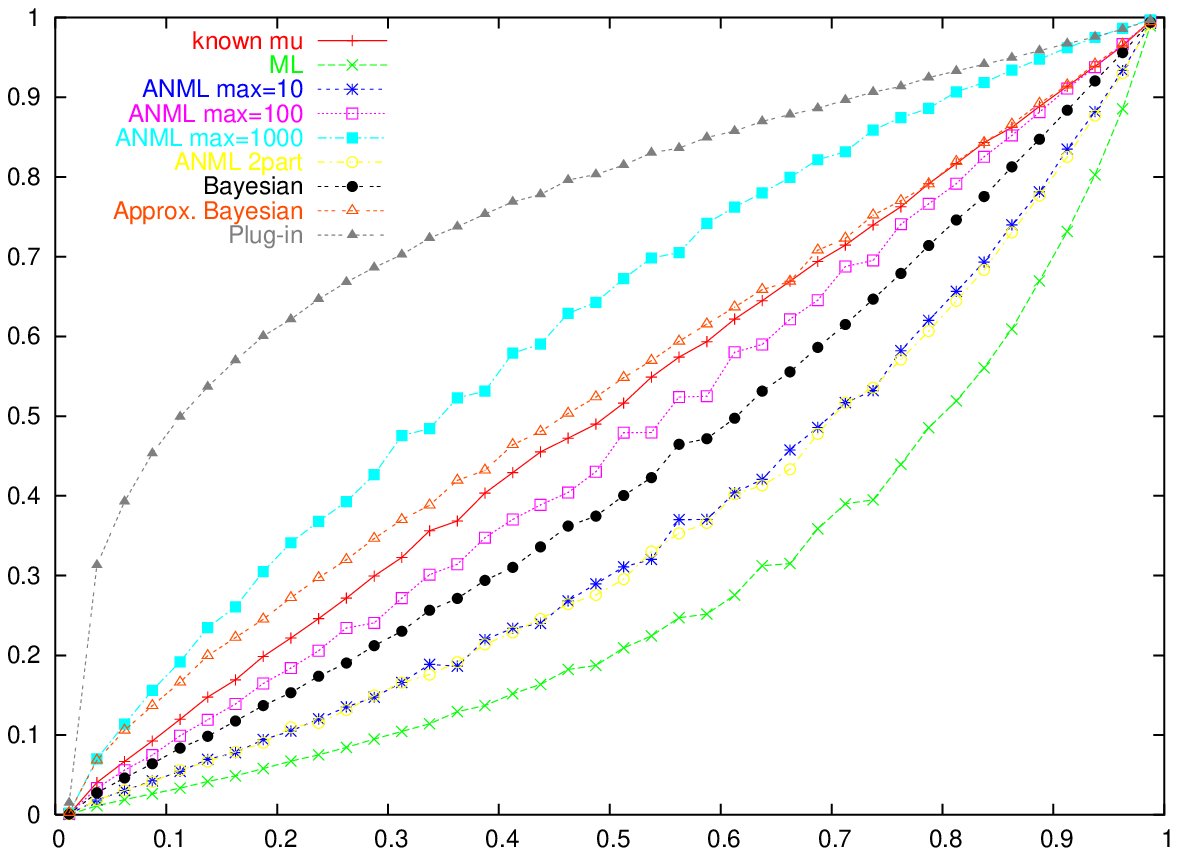}
\end{center}
\caption{\small Calibration: probability that the model assigned to
  the data being Poisson against the frequency with which it actually
  was Poisson. The sample size is $8$ and the mean is $8$.}
\label{fig:calibration}
\end{figure}

\subsection{Calibration}\label{section:cali}
The classical interpretation of probability is frequentist: an event
has probability $p$ if in a repeated experiment the frequency of the
event converges to $p$. This interpretation is no longer really
possible in a Bayesian framework, since prior assumptions often cannot
be tested in a repeated experiment. For this reason, calibration
testing is avoided by some Bayesians who may put forward that it is a
meaningless procedure from a Bayesian perspective.  On the other hand,
we take the position that even with a Bayesian hat on, one would like
one's inference procedure to be calibrated -- in the {\em idealised\/}
case in which identical experiments are performed repeatedly,
probabilities should converge to frequencies. If they do not behave as
we would expect even in this idealised situation, then how can we
trust inferences based on such probabilities in the real world with
all its imperfections? We feel that calibration testing is too
important to ignore, safeguarding against inferences or predictions
that bear little relationship to the real world.  Moreover, in the
so-called ``objective Bayes'' branch of Bayesian statistics, one does
emphasise  Bayesian procedures with good frequentist
behaviour.\cite{Berger2004} At least in restricted contexts
\cite{ClarkeBarron1990,ClarkeBarron1994}, Jeffreys' prior has the
property that the Kullback-Leibler divergence between the true
distribution and the posterior converges to zero quickly, no matter
what the true distribution is.
Consequently, after observing only a
limited number of outcomes, it should already be possible to interpret
the posterior as an almost ``classical'' distribution in the
sense that it can be verified by frequentist experiments
\cite{ClarkeBarron1990}.

In the introduction we have indicated the correspondence between
codelengths and probability. If the universal codelengths for the
different criteria correspond to probabilities that make sense in a
frequentist way, then the Bayesian a posteriori probabilities of the
two models should too. To test this, we generate a sample and compute
the a posteriori probability that it is generated by the Poisson
model, for each of the selection criteria. This probability is
discretised into 40 bins. For each bin we count the number of
sequences that actually were generated by Poisson. If the a posteriori
Bayesian probability that the model is Poisson makes any sense in a
frequentist way, then the result should be a more or less straight
diagonal.

We use mean $8$ and sample size $8$ because on the one hand we want a
large enough sample size that the posterior has converged to something
reasonable, but on the other hand if we choose the sample size even
larger it becomes exceedingly unlikely that a sequence is generated of
which the probability that it is Poisson is estimated near $0.5$, so
we would need to generate an infeasibly large number of samples to get
accurate results.

Our results are in Figure~\ref{fig:calibration}. The ``known $\mu$''
criterion is clearly perfectly calibrated, which makes sense since its
implicit prior distribution on the mean of the generating distribution
puts all probability on the actual mean, so the prior perfectly
reflects the truth in this case. Under such circumstances Bayesian and
frequentist probability become the same, so we get a perfect answer.

Clearly the ML and plug-in criteria, which were observed to be the
most biased in Section~\ref{section:perr}, are again the two worst
performers. When the probability that the data is Poisson distributed
is assessed by the ML criterion to be around $0.5$, the real frequency
of the data being Poisson distributed is only about $0.2$. The plug-in
criterion behaves even worse, assigning a probability of $0.6$ to the
Poisson model for sequences of which only $0.2$ actually were Poisson
distributed. The other criteria are better calibrated.

The calibration test does seem to be very sensitive, exposing
weaknesses of selection criteria much more clearly than error rate
tests. In fact, if circumstances allow it, one of the best ways to
engineer a selection criterion may be to just use the GLRT, optimising
the threshold for performance on a calibration test.

\section{Summary and conclusion}\label{section:conclusion}
We have performed error rate tests and calibration tests to study the
properties of a number of model selection criteria. These criteria are
based on the MDL philosophy and involve computing the code length of
the data with the help of the model. There are several ways to compute
such a code, but the preferred method, the Normalised Maximum
Likelihood (NML) code, cannot be applied since it is not well-defined
for the Poisson and geometric models that we consider.

We have experimented with the following alternative ways of working
around this problem: (1) using BIC which is a simplification of
approximated NML (ANML), (2) ANML with a restricted parameter range,
this range being either fixed or encoded separately, (3) a Bayesian
model using Jeffreys' prior, which is improper for the case at hand
but which can be made proper by conditioning on the first outcome of
the sample, and (4) a plug-in code which always codes the new outcome
using the distribution indexed by the maximum likelihood estimator for
the preceding outcomes.

Both BIC and ANML with a fixed restricted parameter range define a
GLRT test and can be interpreted as methods to choose an appropriate
threshold. BIC implies a neutral threshold, so the criterion will
become biased in favour of the model which is most susceptible to
overfitting. We found that even though both models under consideration
have only one parameter, a GLRT with neutral threshold tends to be
biased in favour of Poisson.  ANML implies a threshold that
counteracts this bias, but for every such threshold value there exists
a corresponding parameter range, so it doesn't provide any more
specific guidance in selecting that threshold. If the parameter range
is separately encoded, this problem is avoided and the resulting
criterion behaves competitively, although it is not calibrated as well
as the Bayesian criterion and the two-part codelength is slightly
redundant.

The Bayesian criterion displays reasonable performance both on the
error rate experiments and the calibration test. The Bayesian
universal codes for the models are not redundant and admit an MDL
interpretation as minimising worst-case codelength in an expected
sense (Section~\ref{sec:objective}).

The plug-in criterion has a bias in favour of the geometric model that
depends strongly on the sample size. As a consequence its error rate
decreases more slowly in the sample size if we put a prior on the
generating distribution that assigns nonzero probability to both
models. This result was surprising to us and has led to a theoretical
analysis of the codelength of the plug-in code in \cite{GrunwaldR05}. 
It turns out
that the regret of the plug-in code does not necessarily grow with
${k\over2}\ln n$ like the NML and Bayesian codes do, if the sample is
not distributed according to any element of the model. We conjecture
that model selection based on plug-in codes continues to behave
suboptimally in more general settings. However, we should note that
there are strong limits to `how bad things can get'. Various results
\cite{BarronRY98} indicate that model selection based on plug-in codes
must eventually select the correct model (if such a model exists),
even when the number of models under consideration is unbounded.

In conclusion, while NML certainly seems a sensible approach to
defining model selection criteria, when it is undefined it is
impossible to minimise the worst case regret. There are many different
methods to deal with this problem, some of which work reasonably well
and some of which work surprisingly badly. The fundamental question
remains: if it is not possible to minimise the worst case regret, then
what exactly \emph{should} we optimise?

\section*{Acknowledgements}
The main idea for this article is not our own, but comes from Aaron D.
Lanterman's text ``Hypothesis Testing for Poisson versus Geometric
Distributions using Stochastic Complexity'' \cite{Lanterman05} which
is a pleasure to read. He deserves much credit.

This work was supported in part by the IST Programme of the European
Community, under the PASCAL Network of Excellence,
IST-2002-506778. This publication only reflects the authors' views.

\clearpage

\bibliography{modselstudy,master,MDL,thisvolume}

\clearpage

\appendix

\section{Proofs}
\label{appendix:proofs}

\begin{proposition}
The parametric complexity of the restricted geometric model subtracted
from the parametric complexity of the restricted Poisson model is
monotonically increasing.
\end{proposition}
\begin{proof}
We have to show that the derivative is nonnegative
everywhere. Consecutive inequalities should be interpreted as having a
``follows from'' relationship.
\begin{eqnarray*}
  0& \le &{d\over dx}\left(\ln\sqrt x - \ln\ln(\sqrt x+\sqrt{x+1})\right)\\
  & = &{1\over 2x}-{1\over\ln(\sqrt x+\sqrt{x+1})}\cdot {1\over
  4x+2+4\sqrt{x}\sqrt{x+1}}\\
  x& \le &\ln(\sqrt{x}+\sqrt{x+1})(2x+1+2\sqrt{x}\sqrt{x+1})\\
  {x\over 4x+1}& \le &\ln(\sqrt{x}+\sqrt{x+1})\\
  {d\over dx}{x\over 4x+1}& \le &{d\over dx}\ln(\sqrt{x}+\sqrt{x+1})\\
  {1\over (4x+1)^2}& \le &{1\over 2\sqrt{x}\sqrt{x+1}}\\
  16x^2+8x+1& \ge &2\sqrt{x}\sqrt{x+1}
\end{eqnarray*}
From convexity of the logarithm, we have
$\ln(x+{1\over2})\ge{1\over2}\ln(x)+{1\over2}\ln(x+1)=\ln(\sqrt{x}\sqrt{x+1})$;
by monotonicity of logarithm it follows that
$x+{1\over2}\ge\sqrt{x}\sqrt{x+1}$; we use this to bound the right
hand side term in the final inequality. Thus the proof follows from
$16x^2+8x+1\ge2x+1$.
\end{proof}

\end{document}